\documentclass[11pt]{article}

\usepackage{fullpage}
\usepackage{amsmath}
\usepackage{amssymb}
\usepackage{color}
\usepackage{authblk}
\DeclareUnicodeCharacter{2113}{$\ell$}
\usepackage{hyperref}
\usepackage{fancyhdr}

\usepackage{float}
\usepackage{graphicx}
\usepackage[font={small,it}]{caption}
\usepackage[font={small,it}]{subcaption}
\usepackage{algorithm}
\usepackage{algpseudocode}

\usepackage{enumitem}

\usepackage[title]{appendix}

\usepackage{array}
\newcolumntype{L}[1]{>{\raggedright\let\newline\\\arraybackslash\hspace{0pt}}m{#1}}
\newcolumntype{C}[1]{>{\centering\let\newline\\\arraybackslash\hspace{0pt}}m{#1}}
\newcolumntype{R}[1]{>{\raggedleft\let\newline\\\arraybackslash\hspace{0pt}}m{#1}}

\newcommand{\gpdf}[2]{\ensuremath{\mathcal{N}\left(\cdot;#1,#2\right)}}
\newcommand{\genpdf}[3]{\ensuremath{\mathcal{GN}\left(\cdot;#1,#2,#3\right)}}
\newcommand{\gpdfX}[3]{\ensuremath{\mathcal{N}\left(#1;#2,#3\right)}}
\newcommand{\expect}[1]{\ensuremath{\mathbb{E}\left[#1\right]}}
\newcommand{\expectX}[2]{\ensuremath{\mathbb{E}_{#1}\left[#2\right]}}
\newcommand{\var}[1]{\ensuremath{\text{Var}\left[ #1 \right]}}
\newcommand{\varX}[2]{\ensuremath{\text{Var}_{#1}\left[ #2 \right]}}

\newcommand{\x}{\ensuremath{\mathbf{x}}}
\newcommand{\y}{\ensuremath{\mathbf{y}}}
\newcommand{\w}{\ensuremath{\boldsymbol{\omega}}}
\newcommand{\nn}[2]{\ensuremath{f^{#1} \left( #2 \right)}}
\newcommand{\kth}[1]{\ensuremath{#1^{[k]}}}

\newcommand{\dist}[1]{\ensuremath{\left\lVert #1 \right\rVert_2^2}}
\newcommand{\dcov}[2]{\ensuremath{\left\lVert #2 \right\rVert_{#1}^2}}

\begin{document}

\title{Empowering Bayesian Neural Networks with Functional Priors through Anchored Ensembling for Mechanics Surrogate Modeling Applications.}
\author[1]{Javad Ghorbanian}
\author[1]{Nicholas Casaprima}
\author[1]{Audrey Olivier}
\affil[1]{Sonny Astani Department of Civil and Environmental Engineering, University of Southern California}
\date{}
\maketitle
\footnotetext[1]{Email addresses: \texttt{mg16979@usc.edu} (Javad Ghorbanian) \texttt{ndearauj@usc.edu} (Nicholas Casaprima) \texttt{audreyol@usc.edu} (Audrey Olivier)}

\thispagestyle{fancy}
\fancyhf{}
\fancyfoot[L]{\textbf{Preprint, Under review.}} 

\abstract{In recent years, neural networks (NNs) have become increasingly popular for surrogate modeling tasks in mechanics and materials modeling applications. While traditional NNs are deterministic functions that rely solely on data to learn the input--output mapping, casting NN training within a Bayesian framework allows to quantify uncertainties, in particular epistemic uncertainties that arise from lack of training data, and to integrate \textit{a priori} knowledge via the Bayesian prior. However, the high dimensionality and non-physicality of the NN parameter space, and the complex relationship between parameters (NN weights) and predicted outputs, renders both prior design and posterior inference challenging. In this work we present a novel BNN training scheme based on anchored ensembling that can integrate \textit{a priori} information available in the function space, from e.g. low-fidelity models. The anchoring scheme makes use of low-rank correlations between NN parameters, learnt from pre-training to realizations of the functional prior. We also perform a study to demonstrate how correlations between NN weights, which are often neglected in existing BNN implementations, is critical to appropriately transfer knowledge between the function-space and parameter-space priors. Performance of our novel BNN algorithm is first studied on a small 1D example to illustrate the algorithm's behavior in both interpolation and extrapolation settings. Then, a thorough assessment is performed on a multi--input--output materials surrogate modeling example, where we demonstrate the algorithm's capabilities both in terms of accuracy and quality of the uncertainty estimation, for both in-distribution and out-of-distribution data.}


\section{Introduction}

In recent years deep learning has become an invaluable tool in mechanics and materials sciences. Neural networks (NNs) can be leveraged as surrogates of computationally expensive physics-based simulators, enabling complex tasks that require many calls to the underlying simulator such as optimization \cite{WHITE2019}, uncertainty quantification \cite{WANG2021,SHAHANE2022}, or multi-scale modeling \cite{Aldakheel2023,KALINA2024}. Beyond surrogate modeling, recent works have also demonstrated the potential of NNs to discover meaningful constitutive models from data \cite{sparse_discovery,BENADY2024}. 

A major drawback of traditional NNs however is their lack of embedded uncertainty quantification (UQ). Quantifying uncertainties in data-driven models, in particular epistemic uncertainties that arise from lack of data, is critical to assess confidence in predictions in small data regimes and extrapolatory conditions. In addition to improving the safety and trustworthiness of machine learning (ML) models, critical for integration within high-consequence decision-making frameworks \cite{Olivier2023}, uncertainty predictions can help optimize data collection procedures \cite{BHOURI2023}, of interest in many engineering fields where experimental or simulation data is expensive to obtain.

The Bayesian framework is routinely used to quantify uncertainties in physics-driven parameters and models, e.g. \cite{Olivier2018,Khodadadian2020,CHAKRABORTY2021,REINER2023} to cite only a few examples in mechanics. In theory, Bayesian neural networks (BNNs), which learn a distribution over the NN parameters instead of a single point estimate, could provide uncertainty estimates on the outputs, alongside a means to integrate \textit{a priori} domain knowledge through the Bayesian prior. In practice however, Bayesian inference of NNs is notoriously challenging due to the high dimensionality and non physicality of the parameter space, and the nontrivial relationship between the parameter and function spaces, which complicates both the formulation of a meaningful prior and inference of the posterior.

With respect to posterior inference, several methods exist that achieve different trade-offs between computational complexity vs. accuracy of the uncertainty estimates. Variational inference (VI) schemes \cite{Blundell2015}, including Monte Carlo dropout \cite{Gal2016}, have been shown to under estimate uncertainties \cite{Olivier2021}, while sampling methods such as the Hamiltonian Monte Carlo (HMC, \cite{Neal2011}) are more accurate but extremely computationally expensive for even moderate size NNs. Other methods that achieve a better trade-off between computational complexity and accuracy include tractable approximate Gaussian inference \cite{tagi}, or deep ensembles \cite{Lakshminarayanan2017}. Vanilla ensembles consist in training an ensemble of NNs starting from different random weight initializations; they can still suffer from under-estimation of uncertainty, and do not integrate information from the prior (thus are not Bayesian methods). Anchored ensembles \cite{Osband2018,Pearce2018} on the other hand encode the Bayesian prior by regularizing each NN to a different sample from the prior distribution, which helps preserve diversity between the NNs and avoid uncertainty collapse. The concept of anchoring was first introduced in \cite{Kirkpatrick2017} for deterministic NNs as a means to mitigate catastrophic forgetting for sequential multi-task learning. When integrated within an ensembling scheme, anchoring has been theoretically linked to concepts of Bayesian inference in e.g. \cite{Pearce2018,Milios2021,Izmailov2021}.

Given a prior density, ensembling with anchoring provides a computationally affordable means to perform approximate posterior inference. A major challenge remains in designing a meaningful prior density in the high-dimensional, non-physical space of NN weights and biases. The choice of prior is particularly critical in low data regimes and extrapolatory conditions as it will greatly impact mean and uncertainty predictions. The important task of designing priors for BNNs is currently receiving attention from both the engineering and machine learning research communities, and can be broadly separated into methods that directly study the impact of different parameter-space priors, and methods that aim to convey information in the function space through functional priors. With respect to the former, recent works \cite{fortuin2022,fortuin2022b} have shown that the widely used isotropic zero-mean Gaussian prior (used for instance in \cite{Olivier2021,LINKA2022,pasparakis2024} in an engineering mechanics setting) is sub-optimal and can yield unintended negative consequences during inference. More importantly, it cannot convey meaningful domain knowledge about the problem at hand.

Since \textit{a priori} information is often available in the function space rather than the parameter space, e.g. through related datasets, low-fidelity models etc., recent works have considered designing priors directly in the function space. In \cite{LINKA2022} and \cite{Meng2022}, functional priors are designed using physics-informed NNs (PINNs) and physics-informed generative adversarial networks (PI-GANs) respectively, then posterior inference is performed via HMC. \cite{Sam2024} presents a VI-based scheme to learn priors that embed general domain knowledge and transfer this learned knowledge across NN architectures. Functional VI \cite{fVI_1,fVI_2} enables specifying priors and performing inference directly in the function space; an important component of these techniques is the design of the measurement set, points at which the functional prior is evaluated, which will impact the relative importance of the prior vs. data. With respect to ensembling techniques, as considered in this work, \cite{YANG2022} proposes randomized prior networks where each prediction in the ensemble consists in a weighted sum of a prior function and a NN fitted to data, while in \cite{Olivier2023} we proposed an ensembling scheme where NNs are concurrently fitted to the data and functional prior samples. Once again, both methods depend on a critical hyper-parameter, chosen empirically, which governs the relative importance of the data vs. functional prior on the posterior predictions. While different methods exist to encode functional prior knowledge within BNNs, the literature lacks an understanding of how these functional priors translate to the parameter-space.

This work aims at bridging the gap between parameter- and function-space priors in BNNs. Indeed, while \textit{a priori} information is typically available in the function space, proper application of Bayes theorem -- which should automatically and appropriately weight data vs. prior information -- does require formulation of a parameter-space prior density. In section \ref{sec:prior_study} we perform a study to uncover important characteristics of parameter-space densities fitted to different functional priors. In particular, we show that, although parameter-space densities exhibit non-Gaussian characteristics, considering correlation between weights is more critical to embed functional knowledge. Building on these insights, we design an enhanced ensembling with anchoring algorithm -- considering low-rank correlation between NN parameters -- to train BNNs given a functional prior. We show that predictions appropriately fit the data while being guided by \textit{a priori} functional and uncertainty information in extrapolatory conditions. We finally apply this algorithm to a materials surrogate modeling problem with out-of-distribution data, providing insights on how to design a functional prior in mechanics problems (section \ref{sec:applications}). The code is available at \url{https://github.com/AudOlivier/BNN_anchored_ensembling}.

\section{Embedding functional priors in BNNs via anchored ensembles with correlated regularization}\label{sec:prior_study}

\subsection{Review of Bayesian neural networks as anchored ensembles}

We consider a regression task that aims at building an input $\x$ to output $\y$ mapping from noisy data $\mathcal{D}=\lbrace (\x_i,\y_i) \rbrace_{i=1:N}$. We represent this unknown mapping via a fully-connected NN $f^{\w}: \x \rightarrow \y$ with $L$ hidden layers, parameterized by its kernels and biases $W^{[l]}, b^{[l]}, l=0:L$, concatenated in parameter vector $\w \in \mathbb{R}^d$. Traditional deterministic NNs learn a point estimate $\hat{\w}$, make deterministic predictions $\y^{\star}=f^{\hat{\w}}\left( \x^{\star} \right)$ and assess accuracy on average over a test set (in-distribution accuracy). Bayesian NNs on the other hand aim at providing a more granular quantification of uncertainties in their predictions. Uncertainties in data-driven modeling are typically categorized as aleatory vs. epistemic \cite{Hullermeier2021}. Aleatory uncertainties arise from the inherent stochasticity or noise in the data, herein we assume a known homoscedastic Gaussian noise model:
\begin{equation}\label{eq:like}
p \left( \mathbf{y} \vert \mathbf{x}, \boldsymbol{\omega} \right) = \gpdfX{\y}{\nn{\w}{\x}}{\Sigma_{\text{noise}}}
\end{equation}
Heteroscedastic, non-Gaussian aleatory noise models can be considered by appropriately modifying the above likelihood function, see for instance \cite{Olivier2023} which considered an input-dependent log-normal likelihood in NN-based travel time prediction applications. 

Epistemic uncertainties arise from lack of training data and are critical in low-data regimes and extrapolatory conditions (out-of-distribution predictions). BNNs quantify epistemic uncertainty by learning a probability distribution over the NN weights, given through Bayes' theorem as:
\begin{equation}
p(\boldsymbol{\omega} \vert \mathcal{D}) \propto p(\mathcal{D} \vert \boldsymbol{\omega}) p_0(\boldsymbol{\omega}) = \left[ \prod_{i=1}^N p \left( \mathbf{y}_i \vert \mathbf{x}_i, \boldsymbol{\omega} \right) \right] p_0(\boldsymbol{\omega})
\end{equation} 
where $p_0(\w)$ is a prior density defined in the high-dimensional parameter space. Posterior uncertainty in the parameter propagates to the output to yield the posterior predictive density:
\begin{equation}
p \left( \y \vert \x, \mathcal{D} \right) = \int p \left( \y \vert \x, \w \right) p(\w \vert \mathcal{D}) d\w
\end{equation}
with mean and variance given as \cite{Olivier2021} (under Gaussian likelihood assumption Eq. \eqref{eq:like}):
\begin{subequations}
\begin{align}
& \expect{ \mathbf{y}^{\star} \vert \mathbf{x}^{\star}, \mathcal{D}} = \expectX{\boldsymbol{\omega} \vert \mathcal{D}}{\expect{\mathbf{y}^{\star} \vert \mathbf{x}^{\star}, \boldsymbol{\omega}}} = \expectX{\boldsymbol{\omega} \vert \mathcal{D}}{f^{\boldsymbol{\omega}}(\mathbf{x}^{\star})} \\
& \var{ \mathbf{y}^{\star} \vert \mathbf{x}^{\star}, \mathcal{D}} = \varX{\boldsymbol{\omega} \vert \mathcal{D}}{\expect{\mathbf{y}^{\star} \vert \mathbf{x}^{\star}, \boldsymbol{\omega}}} + \expectX{\boldsymbol{\omega} \vert \mathcal{D}}{\var{\mathbf{y}^{\star} \vert \mathbf{x}^{\star}, \boldsymbol{\omega}}} = \varX{\boldsymbol{\omega} \vert \mathcal{D}}{f^{\boldsymbol{\omega}}(\mathbf{x}^{\star})} + \Sigma_{\text{noise}}
\end{align} 
\end{subequations}

Vanilla deep ensembles provide an approximation of the epistemic uncertainty by deterministically training an ensemble of $K$ NNs (typically $K \sim 10 - 50$), where each NN is initialized to a different random set of weights, and fitted to a resampled version of the dataset. Resampling of the data can be performed in two ways, via bootstrapping or, if the likelihood model Eq. \eqref{eq:like} is accurately known, via likelihood resampling $\tilde{\y}_i \sim \gpdf{\y_i}{\Sigma_{\text{noise}}}$, $i=1:N$. Algorithm \ref{alg:vanilla} summarizes the training procedure of a vanilla ensemble, where we use the notation $\dcov{S}{\mathbf{z}}=\mathbf{z}^T S^{-1} \mathbf{z}$ in the loss function, which arises from the mutlivariate likelihood Eq. \eqref{eq:like}. Posterior predictive mean and variance are computed as averages over the trained ensemble:
\begin{subequations}\label{eq:predictive_moments_ens}
\begin{align}
& \expect{ \y^{\star} \vert \x^{\star}, \mathcal{D}} = \frac{1}{K} \sum_{k=1}^K \nn{\kth{\hat{\w}}}{\x^{\star}} \\
& \var{ \y^{\star} \vert \x^{\star}, \mathcal{D}} = \frac{1}{K-1} \sum_{k=1}^K \left( \nn{\kth{\hat{\w}}}{\x^{\star}} - \expect{ \y^{\star} \vert \x^{\star}, \mathcal{D}} \right) \left( \nn{\kth{\hat{\w}}}{\x^{\star}} - \expect{ \y^{\star} \vert \x^{\star}, \mathcal{D}} \right)^T + \Sigma_{\text{noise}}
\end{align}
\end{subequations}

\begin{algorithm*}[t!]
\begin{algorithmic}
\State{\textbf{Input}: Data $\mathcal{D}=\lbrace (\x_{i}, \y_{i}) \rbrace_{i=1:N}$, Gaussian likelihood Eq. \eqref{eq:like}}
\State{\textbf{Output}: Ensemble of NNs with weights $\kth{\hat{\w}}$, $k=1:K$, fitted to data $\mathcal{D}$}
\For {$k=1:K$}
\State Resample dataset: $\tilde{\mathcal{D}}=\lbrace (\tilde{\x}_{i}, \tilde{\y}_{i}) \rbrace_{i=1:N} = \text{resample} \left( \mathcal{D}\right)$
\State Randomly initialize NN weights
\State Train NN by minimizing negative log-likelihood: 
$$\kth{\hat{\w}} = \text{argmin} \quad \sum_{i=1}^N \dcov{\Sigma_\text{noise}}{\tilde{\y}_{i} - \nn{\w}{\tilde{\x}_{i}}} $$
\EndFor
\end{algorithmic}
\caption{Vanilla neural network ensembling}\label{alg:vanilla}
\end{algorithm*}

Vanilla ensembles are prone to under-estimate uncertainties, as will be illustrated in the examples, and do not allow consideration of a prior density. To embed \textit{a priori} knowledge from a parameter space density $p_0(\w)$, anchored ensembles regularize each NN to a different sample from the prior $\kth{\w_0} \sim p_0(\cdot)$, that is, the $k^{\text{th}}$ NN in Alg. \ref{alg:vanilla} is trained as:
\begin{align}\label{eq:loss_anchoring_factorized}
\kth{\hat{\w}} &= \text{argmin} \lbrace \log{p(\mathcal{D} \vert \w)} + \log{p_0(\w-\kth{\w_0})} \rbrace \\
 &= \text{argmin} \lbrace \quad \underbrace{\sum_{i=1}^N \dcov{\Sigma_\text{noise}}{\tilde{\y}_{i} - \nn{\w}{\tilde{\x}_{i}}}}_{\text{data fit (likelihood)}} + \underbrace{\lambda \dist{\w-\kth{\w_0}}}_{\text{regularization (prior)}} \rbrace
\end{align}
The above optimization is equivalent to \textit{maximum a posteriori} estimation under the assumption of an isotropic Gaussian prior over the NN weights with precision (inverse of the variance) $\lambda$. Ensembling with anchoring is known to provide consistent estimates of the posterior in the Gaussian linear case \cite{Osband2018}, and \cite{Milios2021} showed that such ensemble learning yields improvement of the divergence between the approximate and the actual posterior for DNNs with ReLU activation functions. Training of anchored ensembles can be parallelized and utilize existing optimization frameworks for training of deterministic NNs, thus they provide a computationally affordable and scalable alternative for approximating BNNs. 

As previously mentioned, assuming an isotropic Gaussian prior is sub-optimal \cite{fortuin2022,fortuin2022b}, it also requires careful tuning of the parameter variance term $1/\lambda$ as the propagation of uncertainties from the parameter to the output space depends on the NN architecture. Most importantly, such simple parameter space prior does not allow to encode \textit{a priori} knowledge typically available in the function space, via e.g. low-fidelity models, related datasets etc. In order to encode such functional knowledge, an enhanced regularization scheme that accounts for correlation between NN parameters is needed, as demonstrated in the following section. 




\subsection{Understanding the relationship between parameter and function-space priors}

Bayes inference is widely used to estimate parameters in physics-driven models, where parameters consist in physical quantities for which \textit{a priori} information or physical constraints may be available to guide the definition of the parameter-space prior $p_0(\cdot)$. In NN models on the other hand, the parameter space is non-physical and high-dimensional, hindering the formulation of the parameter-space prior required to explicitly apply Bayes theorem. Instead, \textit{a priori} information is typically available about the resultant mapping $x \rightarrow y$, via e.g. low fidelity models. In this section we study the connection between parameter and function space priors, to discover which characteristics (non-Gaussianity, anisotropy, correlation) of the parameter-space prior are required to appropriately encode information originally available in the function space. In particular, we demonstrate that accounting for anisotropy and (partial) correlation between parameters is critical, and that it can be done efficiently within an ensembling framework.



\subsubsection{Mapping a functional prior to the parameter space}

We assume access to a functional prior density $p(g)$, where $g: \x \rightarrow \y$, from which we can draw realizations $\kth{g}$, $k=1:K$. For instance, Gaussian random processes (GPs) or low fidelity models with randomized parameters can serve as functional priors. We can then fit an ensemble of $K$ NNs to these realizations of the functional prior, yielding pre-trained weights $\kth{\w_0}$, $k=1:K$, and study the characteristics of the resulting distribution over the NN parameters $p_0(\w) \approx \frac{1}{K} \sum_{k=1}^K \delta(\w - \kth{\w_0})$. Additional details about the (pre-)training procedure are given in Appendix \ref{app:pretraining}.

\begin{figure}[tb!]
\begin{center}
\includegraphics[width=\textwidth]{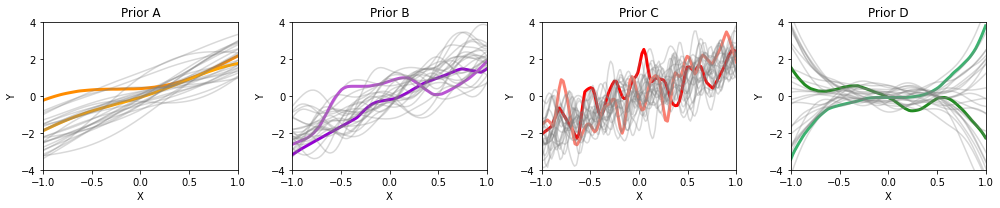}
\caption{Functional priors considered for the function- to parameter-space prior study. Each gray curve represents a realization $\kth{g}$ from the functional prior. The colored lines represent two NNs fitted to two specific realizations.}\label{fig:priors_true}
\end{center}
\end{figure}

For purpose of comparison, in this study we consider 4 GP-based priors, illustrated in Fig. \ref{fig:priors_true}, namely:
\begin{itemize}
\item prior A: a GP with mean $m(x)=2x$ and RBF kernel $k(x,x')=k_0 \exp{\left(-\frac{(x-x')^2}{2L^2}\right)}$ with $k_0=0.6, L=0.8$,
\item prior B: same as prior A with $L=0.2$,
\item prior C: same as prior A with $L=0.05$,
\item prior D: a GP with random mean $m(x)=5 u x^3$ where $u$ follows a uniform distribution $u \sim U(-1,1)$, and RBF kernel with $k_0=0.1, L=0.2$.
\end{itemize}
Priors A to C have the same mean function, which embeds \textit{a priori} knowledge that the output increases with the input, and same marginal densities, but different correlation lengths, yielding functions that evolve more or less rapidly with the input $\x$. Prior A which exhibits the strongest correlation between neighboring points can be seen as a very informative prior, while prior C is more flexible and thus less informative. Prior D overall has zero-mean but exhibit uncertainty that is not constant over the whole domain.

Four ensembles of $K=100$ NNs each (1 input, 1 output, 4 hidden layers with 50 units each resulting in 7801 trainable parameters) are fitted to realizations from these functional priors following the procedure described in Appendix \ref{app:pretraining}. In particular we insist on the fact that the NNs in an ensemble are not randomly initialized, instead they are initialized as slightly perturbed versions of a common set of weights. Starting from the same set of weights is critical to ensure that variations between NN weights after pre-training are representative of training to different prior realizations rather than different initializations, and in particular is necessary to capture meaningful correlations between weights. For each ensemble, representing a different prior, we compute statistics of the pre-trained weights to draw insights on their probability distribution. For instance, the non-zero parameter-prior mean $\mu_0$, needed to encode non-zero prior predictive distributions, is learnt by averaging over the $K$ instances in the ensemble: 
\begin{equation}\label{eq:w_prior_mean}
\mu_0 = \expectX{\text{ens}}{\w} = \frac{1}{K} \sum_{k=1}^K \kth{\w_0}
\end{equation}
Once again, we point out that this average is only meaningful if all the NNs in the ensemble are initialized with the same set of weights.

We first study characteristics of the marginal densities of the demeaned weights $\w_j-\mu_{0,j}$, $j=1:d$, assuming that weights within a given layer follow the same marginal distribution. Table \ref{tab:prior_variance} shows that the variance of the pre-trained NN weights increases when trained to more flexible functions (from prior A to C), consistent with the idea that a stronger regularization $\lambda \propto \frac{1}{\sigma_0^2}$ is needed to enforce a highly informative prior (prior A). Table \ref{tab:prior_variance} also shows significant variations in the variance of kernels in different layers, that is, the prior distribution obtained via pre-training is non-isotropic.

\begin{table}[H]
\centering
\begin{tabular}{c || c || c | c | c | c | c}
parameter set & $\sigma_{0}^2$ & $W^{(0)}$ & $W^{(1)}$ & $W^{(2)}$ & $W^{(3)}$ & $W^{(4)}$ \\ \hline \hline
prior A & $1.8e^{-3}$ & $0.8e^{-3}$ & $2.0e^{-3}$ & $2.0e^{-3}$ & $1.6e^{-3}$ & $1.0e^{-3}$ \\ \hline
prior B & $8.7e^{-3}$ & $2.1e^{-3}$ & $7.6e^{-3}$ & $8.9e^{-3}$ & $10e^{-3}$ & $8.0e^{-3}$ \\ \hline
prior C & $67e^{-3}$ & $9.7e^{-3}$ & $57e^{-3}$ & $59e^{-3}$ & $88e^{-3}$ & $132e^{-3}$ \\ \hline
prior D & $8.5e^{-3}$ & $2.1e^{-3}$ & $9.1e^{-3}$ & $8.3e^{-3}$ & $8.4e^{-3}$ & $6.5e^{-3}$
\end{tabular}
\caption{Variance of kernels pre-trained to different functional priors, on average over the entire NN ($\sigma_{0}^2$), and per layer $W^{(l)}, l=0:4$.\label{tab:prior_variance}}
\end{table}

\begin{figure}[tb!]
\begin{subfigure}[b]{\textwidth}
\centering
\includegraphics[width=\textwidth]{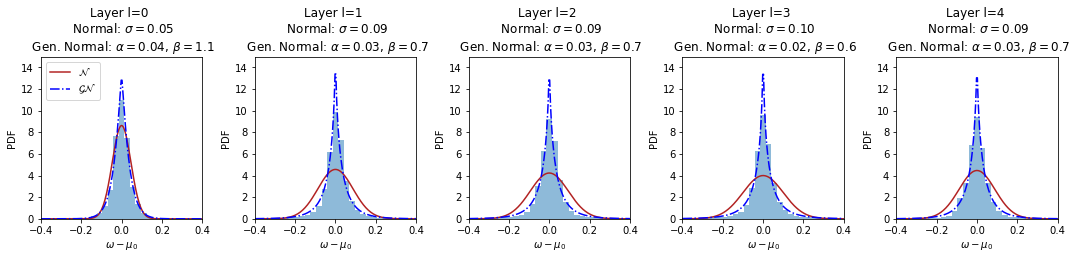}
\caption{Prior B}
\end{subfigure}
\begin{subfigure}[b]{\textwidth}
\centering
\includegraphics[width=\textwidth]{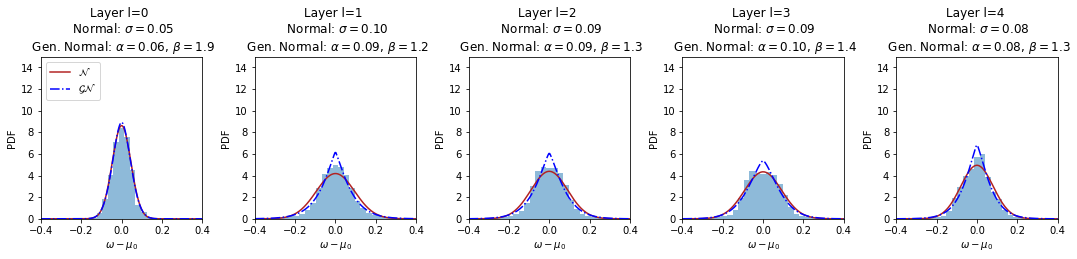}
\caption{Prior D}
\end{subfigure}
\caption{Histogram of NN weights (kernels) pre-trained to priors B and D.\label{fig:prior_dists}}
\end{figure}

Fig. \ref{fig:prior_dists} shows the marginal distributions of the demeaned weights $\w-\mu_{0}$ for priors B and D, which exhibit non-Gaussian behavior, in particular heavy-tails. Gaussian and generalized normal fits are compared to quantify this high kurtosis, where the generalized normal density with mean $\mu_0$ has with PDF $p_0(\omega) \propto \text{exp}\left[ - \left(\vert \omega - \mu_0 \vert / \alpha_0 \right)^{\beta_0} \right]$ ($\beta=2$ recovers a Gaussian distribution, equivalent to l2 regularization, and $\beta=1$ recovers a Laplace distribution, equivalent to l1 regularization). These fits display values of shape parameter $\beta_0$ less than 2 throughout the NNs for all priors, again highlighting the non-Gaussian characteristic of parameter-space priors.

\begin{figure}[tb!]
\begin{subfigure}[b]{\textwidth}
\centering
\includegraphics[width=\textwidth]{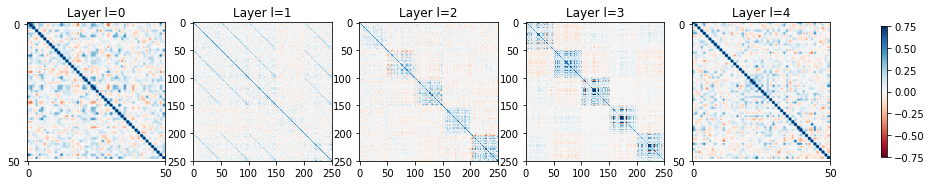}
\caption{Prior B}
\end{subfigure}
\begin{subfigure}[b]{\textwidth}
\centering
\includegraphics[width=\textwidth]{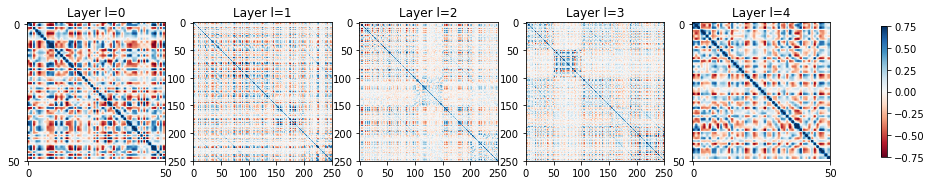}
\caption{Prior D}
\end{subfigure}
\caption{Correlation between some of the NN weights pre-trained to priors B and D.\label{fig:prior_corr}}
\end{figure}

Beyond the marginal distributions, Fig. \ref{fig:prior_corr} displays approximate correlations between some of the NN weights for priors B and D, highlighting the fact that training to different priors induces varying patterns of non-zero correlations between trained weights. This implies that a factorized prior may be inappropriate, as confirmed in the next section. Importantly, this correlation matrix is low-rank since it is evaluated from $K$ samples (number of NNs in an ensemble), which will facilitate computations later on.

Based on this analysis, the parameter-space density resulting from pre-training an ensemble of NNs to a functional prior density exhibits anisotropy, correlations, and non-Gaussianity. In the next section, we demonstrate that appropriately capturing the (low-rank) covariance between the weights is critical, more so than capturing non-Gaussianity, supporting the use of a (degenerate) multivariate Gaussian as parameter-space prior $p_0(\w)$.

\subsubsection{Back to the function space: importance of weights correlations}

In the previous section we mapped a functional prior to a parameter-space distribution through pre-training of an ensemble of NNs. Herein we do the reverse mapping, sampling weights $\kth{\w_0}$, $k=1:K$ from different distributions, exhibiting different characteristics such as non-zero mean, non-Gaussianity, correlation etc., and examine the resulting predictive distributions given by sample functions $f^{\kth{\w_0}}: \x \rightarrow \y$.

\begin{figure}[tb!]
\begin{subfigure}[b]{\textwidth}
\centering
\includegraphics[width=\textwidth]{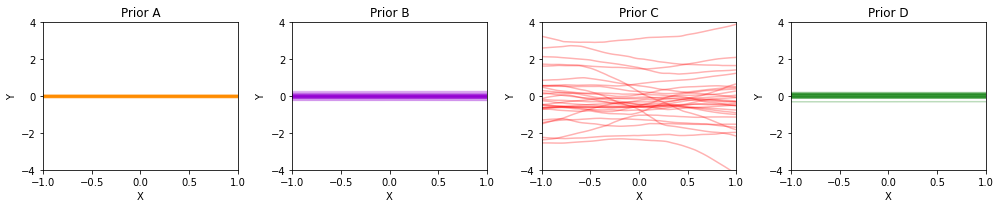}
\caption{}
\end{subfigure}
\begin{subfigure}[b]{\textwidth}
\centering
\includegraphics[width=\textwidth]{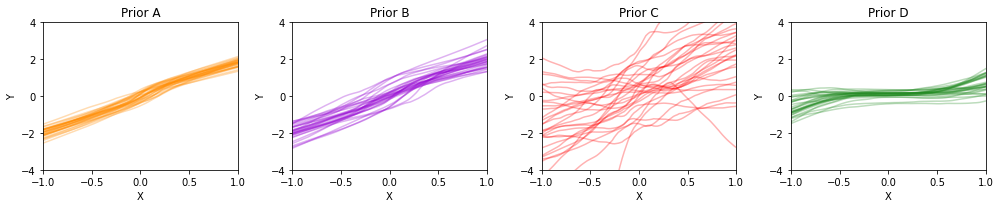}
\caption{}
\end{subfigure}
\begin{subfigure}[b]{\textwidth}
\centering
\includegraphics[width=\textwidth]{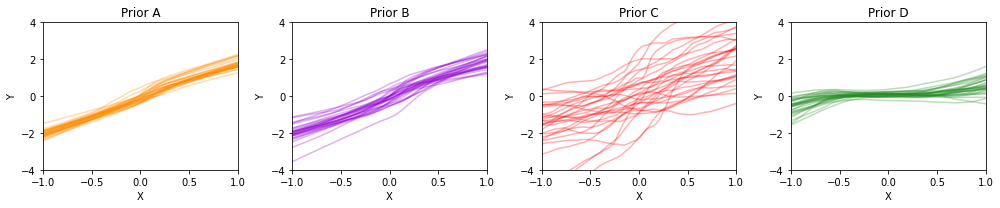}
\caption{}
\end{subfigure}
\begin{subfigure}[b]{\textwidth}
\centering
\includegraphics[width=\textwidth]{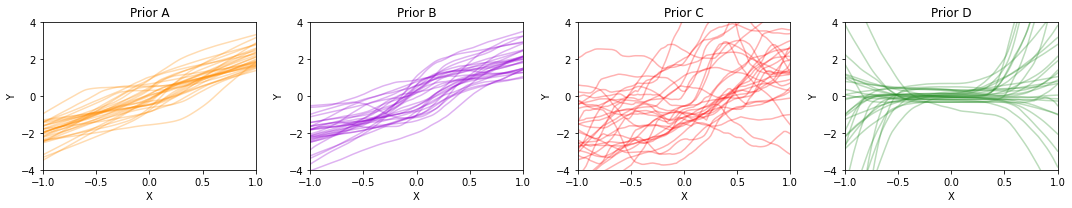}
\caption{}
\end{subfigure}
\caption{Reconstructing functional priors different parameter-space prior distributions: a) zero-mean isotropic Gaussian, b) non-zero mean factorized Gaussian, c) non-zero mean factorized generalized normal, d) multivariate-Gaussian.\label{fig:prior_sampling}}
\end{figure}

Fig. \ref{fig:prior_sampling}~(a) shows functions obtained by sampling NN weights and biases from a traditional zero-mean isotropic Gaussian prior $p_0(\w)=\gpdfX{\w}{0}{\sigma_0^2 \mathbf{I}_d}$ for different values of $\sigma_{0}^2$ (from Table \ref{tab:prior_variance}). While increasing the parameters prior variance yields higher variance predictive distributions, zero-mean weights translate to zero-mean prior predictive distributions and thus do not recover the functional priors.

To account for anisotropy and non-zero mean, we then sample from factorized Gaussian priors $p_0(\w)= \prod_j \gpdf{\mu_{0,j}}{\sigma_{0,j}^2} $ where the variance $\sigma_{0,j}^2$ is shared across weights within a given layer (values per layer reported in Table \ref{tab:prior_variance}). Fig. \ref{fig:prior_sampling}~(b) shows that the mean trend is recovered in the function space, and that the predictive uncertainty somewhat captures higher levels of flexibility from prior A to C. However, this factorized parameter-space prior is not capable of capturing the more complex varying levels of uncertainty in prior D. Very similar conclusions can be reached when replacing the factorized Gaussian with a factorized generalized normal $p_0(\w)= \prod_j \genpdf{\mu_{0,j}}{\alpha_{0,j}}{\beta_{0,j}} $ (Fig. \ref{fig:prior_sampling}~(c)), which seems to indicate that non-Gaussianity is not the most critical parameter-space characteristic needed to encode functional knowledge.

Finally, sampling NN parameters from a multivariate Gaussian $p_0(\w)=\gpdfX{\w}{\mu_{0}}{\Sigma_0}$, where $\mu_{0}, \Sigma_0$ are the sample mean and sample covariance from the NN ensemble pre-trained to the functional prior, helps recover, albeit not perfectly, the correct mean and uncertainty in the function space (Fig. \ref{fig:prior_sampling}~(d)). This finding supports the idea that capturing correlations between weights is critical, more so than capturing non-Gaussianity, to convey functional prior knowledge. Very importantly, the  covariance matrix $\Sigma_0$ is low-rank, which enables computation of its inverse in an efficient way even though $\w$ is very high-dimensional, as explained next.

\subsubsection{Exploiting low-rank approximation of the parameter correlation matrix}

Within the proposed framework which relies on ensembling, the (prior) weights covariance matrix $\Sigma_0 \in \mathbb{R}^{d \times d}$ is computed as a sample covariance from $K$ samples $\kth{\w}_0$ (pre-trained weights). Because $K << d$, the covariance matrix is low-rank and can be computed efficiently from a singular value decomposition (SVD) of the centered `data' matrix $\mathbb{W}_0=\left[ \w_0^{[1]}-\mu_0, \cdots, \w_0^{[K]}-\mu_0 \right]^T \in \mathbb{R}^{K \times d}$ (with $\mu_0$ as in Eq. \eqref{eq:w_prior_mean}):
\begin{align}
&\mathbb{W}_0 = U S V^T \quad \text{SVD of centered data matrix} \nonumber \\
&\Sigma_0 = \frac{1}{K-1}  \mathbb{W}_0^T \mathbb{W}_0 = V \frac{S^2}{K-1} V^T
\end{align}
Since $\Sigma_0$ is low-rank, the corresponding multivariate distribution in dimension $d$ is degenerate, with PDF $p_0(\omega) \propto \text{exp}\left( - \frac{1}{2} (\w-\mu_0)^T \Sigma_0^{+} (\w-\mu_0) \right)$ where $\Sigma_0^{+}$ is the generalized-inverse of $\Sigma_0$, which can be computed as (Appendix \ref{app:svd}):
\begin{equation}
\Sigma_0^{+}=(K-1)VS^{-2}V^T
\end{equation}
yielding:
\begin{align}
\ln{p_0(\w)} &= \text{constant} - \frac{K-1}{2} (\w-\mu_0)^T V S^{-2} V^T (\w-\mu_0) \nonumber \\ 
&= \text{constant} - \frac{K-1}{2} \dcov{S^2}{V^T (\w-\mu_0) }
\end{align}
These computations will be leveraged later on to derive an efficient anchoring scheme for BNNs that accounts for correlation between NN weights.

\begin{figure}[tb!]
\centering
\includegraphics[width=0.4\textwidth]{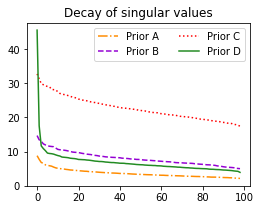}
\caption{Decay of singular values of $\mathbb{W}_0$, matrix of NN weights pre-trained to different priors.\label{fig:prior_decay}}
\end{figure}

In the SVD above, $S$ has only $K-1 << d$ non-zero singular values, and the decay of these singular values can shed light on the NN ensemble size required to capture important trends in the correlation matrix. Fig. \ref{fig:prior_decay} shows the decay of singular values for the four priors considered in this study. Singular values of flexible priors (prior C) overall have larger magnitudes, consistent with flexible priors exhibiting larger parameters variance. Interestingly, for prior D the decay of singular values is very rapid, indicating that even a small-size ensemble could capture the important correlation trend necessary to approximately encode the varying uncertainty pattern embedded in this functional prior.

\subsection{Novel BNN training algorithm: ensembling with correlated regularization}

Based on the insights drawn in the previously detailed study we derive a novel algorithm for training BNNs that can convey \textit{a priori} information (both in terms of mean and uncertainty) available in the function space, as detailed in Alg. \ref{alg:ours}. In particular, the previous findings support the use of a non-zero multivariate Gaussian as parameter-space prior $p_0(\w)=\gpdfX{\w}{\mu_{0}}{\Sigma_0}$ (rather than the widely used zero-mean isotropic Gaussian prior), where $\mu_{0}, \Sigma_0$ are learnt from the NN ensemble pre-trained to the functional prior (pre-training stage). NNs in the ensemble are then fitted to the dataset via \textit{maximum a posteriori}, where the loss includes a data fit term and a regularization term arising from the multivariate Gaussian prior $p_0(\w)$. On condition that the noise variance $\Sigma_{\text{noise}}$ be known, the algorithm does not require fine-tuning of a regularization strength parameter $\lambda$. Instead, appropriate weighting of data vs. prior is embedded in the estimation of the covariance $\Sigma_0$. 

\begin{algorithm*}[t!]
\begin{algorithmic}
\State{\textbf{Input}: Data $\mathcal{D}=\lbrace (\x_{i}, \y_{i}) \rbrace_{i=1:N}$, Gaussian likelihood Eq. \eqref{eq:like}, functional prior $p(g)$, $g: \x \rightarrow \y$}
\State{\textbf{Output}: Ensemble of NNs fitted to data $\mathcal{D}$, embedded with \textit{a priori} knowledge from prior $p(g)$}
\State
\State \underline{Stage 1: Pre-train to functional prior and learn corresponding parameter-space prior}
\State Generate one set of weights $\w_{\text{init}}$ via e.g. He initialization.
\For {$k=1:K$}
\State Sample realization from functional prior $\kth{g} \sim p(g)$
\State Initialize NN weights to a slightly perturbed version of $\w_{\text{init}}$
\State Pre-train NN to fit $\kth{g}$ (see Alg. \ref{alg:pre-training}), yielding pre-trained weights $\kth{\w_0}$
\EndFor
\State Build centered `data' matrix $\mathbb{W}_0=\left[ \w_0^{[1]}-\mu_0, \cdots, \w_0^{[K]}-\mu_0 \right]^T$ with $\mu_0$ sample mean
\State Extract $S, V$ from SVD decomposition $\mathbb{W}_0=USV^T$
\State
\State \underline{Stage 2: Train to data, while regularizing to parameter-space prior}
\For {$k=1:K$}
\State Resample data: $\tilde{\mathcal{D}}=\lbrace (\tilde{\x}_{i}, \tilde{\y}_{i}) \rbrace_{i=1:N} = \text{resample} \left( \mathcal{D}\right)$
\State Fit NN to resampled dataset, regularizing to pre-trained weight $\kth{\w_0}$:
\begin{align*}
\kth{\hat{\w}} &= \text{argmin} \lbrace \quad \sum_{i=1}^N \dcov{\Sigma_\text{noise}}{\tilde{\y}_{i} - \nn{\w}{\tilde{\x}_{i}}} + \dcov{\Sigma_0}{\w-\kth{\w_0}} \rbrace \\
&= \text{argmin} \lbrace \quad \sum_{i=1}^N \dcov{\Sigma_\text{noise}}{\tilde{\y}_{i} - \nn{\w}{\tilde{\x}_{i}}} + (K-1) \dcov{S^2}{V^T \left(\w-\kth{\w_0}\right)} \rbrace
\end{align*}
\EndFor
\end{algorithmic}
\caption{BNN via anchored ensembling with correlated regularization}\label{alg:ours}
\end{algorithm*}

In the next section we study the capability of this algorithm to appropriately fit available data while quantifying epistemic uncertainties, guided by the prior, away from training data. We first consider a simple 1D problem to assess the algorithm's ability to quantify uncertainties in both extrapolation and interpolation. We then apply the algorithm to probabilistic surrogate materials modeling.

\section{Performance of the novel BNN training algorithm}\label{sec:applications}

\subsection{1D example}\label{sec:algo_1d_example}

We consider a 1D mapping with data generated from a sinusoidal function with an increasing mean trend and Gaussian noise. The input is sampled in two disconnected regions to create a scenario where both interpolation and extrapolation can be studied. More specifically, 30 data points $\lbrace (x_{i}, y_{i}) \rbrace_{i=1:30}$ are sampled as follows:
\begin{subequations}
\begin{align}
&x_i \sim U(-0.6, 0.1), \quad i=1:25 \quad \text{and} \quad x_i \sim U(0.6, 0.65), \quad i=26:30 \\
&y_i =  1.5 (x_i-0.2) + \sin{ \left[ 8(x_i-0.2) \right]} + \varepsilon_i, \quad \varepsilon_i \sim \gpdf{0}{0.1^2}
\end{align}
\end{subequations}

\begin{figure}[tb!]
\begin{subfigure}[c]{0.35\textwidth}
\centering
\includegraphics[width=\textwidth]{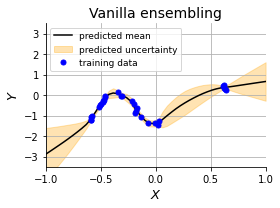}
\caption{}
\end{subfigure}%
\begin{subfigure}[c]{0.65\textwidth}
\centering
\includegraphics[width=\textwidth]{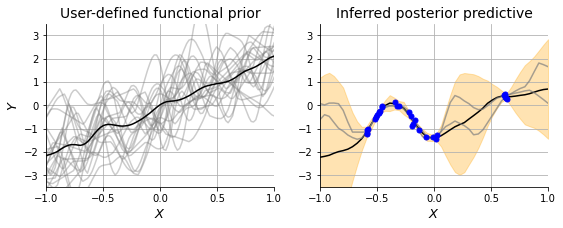}
\includegraphics[width=\textwidth]{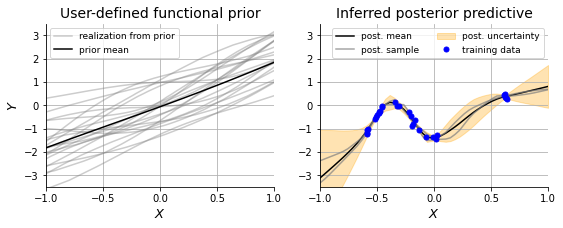}
\caption{}
\end{subfigure}
\caption{1D example: mean and epistemic uncertainty prediction from (a) a vanilla ensemble of $K=40$ NNs and (b) an anchored ensemble (proposed approach, $K=40$) given two different function-space priors. \label{fig:example1D}}
\end{figure}

We represent the input--output mapping with a fully-connected NN with 4 hidden layers, 20 neurons in each layer, and leaky ReLU activation functions with small negative slope $\alpha=0.01$. Fig. \ref{fig:example1D}~(a) shows the predictions of a vanilla ensemble of $K=40$ NNs (Alg. \ref{alg:vanilla}, leveraging a likelihood resampling scheme to resample the data). The mean and epistemic variance predictions are computed by sample averages over the ensemble as in Eq. \eqref{eq:predictive_moments_ens}, without the aleatory variance term $\Sigma_{\text{noise}}$. Throughout the manuscript, reported uncertainties relate solely to the epistemic uncertainty which is plotted as mean $\pm$ 2 standard deviations. The vanilla ensemble, where diversity between individual NNs arises from the data resampling scheme and random parameter initialization, predicts small amounts of uncertainty both in interpolation and extrapolation. Importantly, this uncertainty is under-estimated compared to anchored ensembles studied next, even those considering constrained functional priors.

In a rigorous Bayesian framework, the prior density -- herein provided in the function-space -- can help guide posterior estimation away from data. In this simple example we consider a GP-based prior with a mean that captures the increasing linear trend in the data. Two GPs are considered with lengthscales $L=0.1$ (very flexible prior) and $L=1$ (constrained prior) respectively. Fig. \ref{fig:example1D}~(b) shows that the proposed algorithm (Alg. \ref{alg:ours}) appropriately predicts high levels of uncertainty, even in interpolation, when allowing for complex functions \textit{a priori}. On the other hand, when \textit{a priori} constrained to slowly varying functions, the algorithm does not allow for flexible interpolation between the two data regions and thus appropriately predicts a smaller posterior uncertainty.

\begin{figure}[tb!]
\centering
\includegraphics[width=0.7\textwidth]{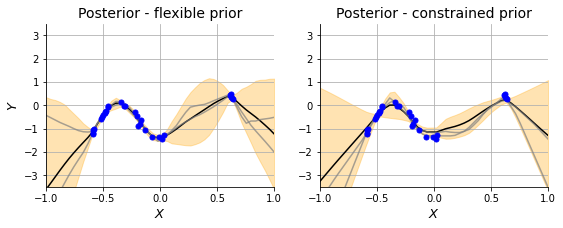}
\caption{1D example: anchored ensembles that neglect correlation between weights (factorized Gaussian prior).\label{fig:example1D_bis}}
\end{figure}

Finally, Fig. \ref{fig:example1D_bis} shows posterior predictions using anchored ensembles that neglect correlation between NN weights (that is, the parameter-space prior is assumed to be a factorized Gaussian density $p_0(\w)=\prod_j \gpdfX{\w_j}{\mu_{0,j}}{\sigma_{0,j}^2}$, inducing a loss as in Eq. \eqref{eq:loss_anchoring_factorized}). Compared to our proposed algorithm which accounts for correlation between NN weights (Fig. \ref{fig:example1D}~(b)), this approach yields an imperfect fit to data with the constrained prior, an under-estimation of uncertainties with the flexible prior, and overall a tendency to forget the prior mean (for values of $x$ close to 1.). This inability to appropriately convey prior information using a factorized parameter-space prior agrees with results reported in e.g. \cite{Olivier2023}, and again highlights the importance of capturing (low-rank) correlations between NN weights within the anchoring scheme.

\subsection{Materials modeling example}\label{sec:materials}

\subsubsection{Problem statement and data generation}

\begin{figure}[bt!]
\begin{center}
\includegraphics[width=0.9\textwidth]{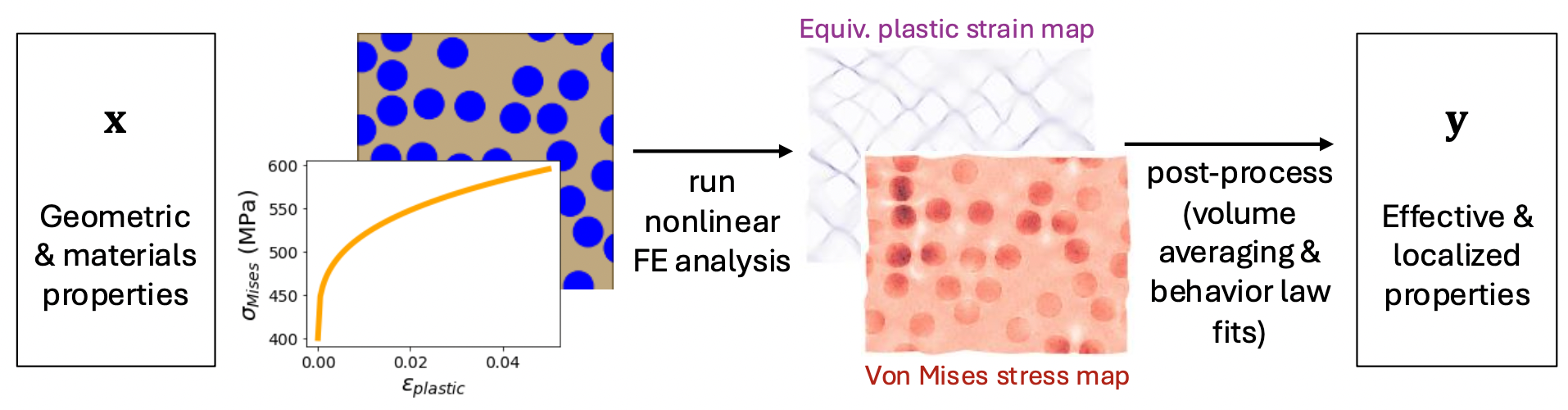}
\end{center}
\caption{Materials surrogate modeling: overview of data generation procedure. \label{fig:materials_overview}}
\end{figure}

A cornerstone task in materials modeling and discovery consists in building efficient structure--property linkages from experimental or simulation data, typically expensive to obtain. Embedding these data-driven linkages with quantification of epistemic uncertainty is critical to assess confidence in predictions and guide future data collection. Herein we consider a surrogate modeling task that maps geometric and materials properties of a two-phase composite microstructure (input $\x$) to effective properties of the representative volume element (output $\y$). In the following we provide a brief introduction to the data generation process, summarized in Fig. \ref{fig:materials_overview}; the reader is referred to \cite{Olivier2021} for a more detailed presentation.

The materials system under study is a 2-phase composite (volume fraction $\text{vf}$) with stiff fibers (Young's modulus $E_f$, 30 circular inclusions) randomly placed in a matrix that exhibits plastic behavior, governed by a power law $\sigma=a + b_m (\varepsilon_{pl})^c_m$ with $a=400$ MPa. Given a set of input parameters $\x=\left[ \text{vf}, E_f, b_m, c_m \right]^T$, we generate a sample microstructure, assign materials properties and run a finite element analysis, applying uni-axial tensile strain in the horizontal direction, under plane strain and periodic boundary conditions. The stress and strain maps are then post-processed to extract effective properties of the representative volume element (RVE): fitting Hooke's law to volume averaged stress/strain before observing any plasticity yields the effective Young's modulus $E_{eff}$ and Poisson ratio $\nu_eff$ of the RVE, while fitting a power law $\sigma_{\text{Mises}}=a + b_{eff} (\varepsilon_{pl})^c_{eff}$ to volume averaged Von Mises/equivalent plastic strain yields the effective plasticity coefficients $b_{eff},c_{eff}$. In addition to homogenized properties, which can be leveraged in e.g. multi-scale simulations, localized properties such as high values of stress in the microstructure can be of interest when studying localized failure mechanisms. We then also extract from the Von Mises stress map the proportion of fibers exhibiting stress above a certain threshold (here $0.9$ GPa), denoted $p_{\sigma}$. Thus the complete output vector of interest is $\y=\left[b_{eff},c_{eff},E_{eff},\nu_{eff},p_{\sigma} \right]^T$.

\begin{figure}[bt!]
\begin{subfigure}[b]{\textwidth}
\centering
\includegraphics[width=0.8\textwidth]{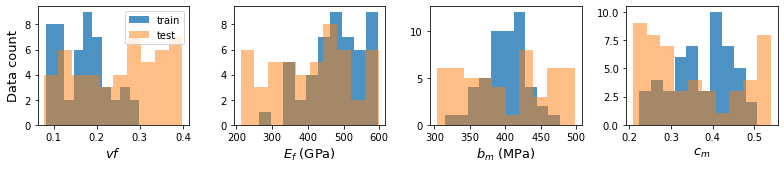}
\caption{}
\end{subfigure}
\begin{subfigure}[b]{\textwidth}
\centering
\includegraphics[width=\textwidth]{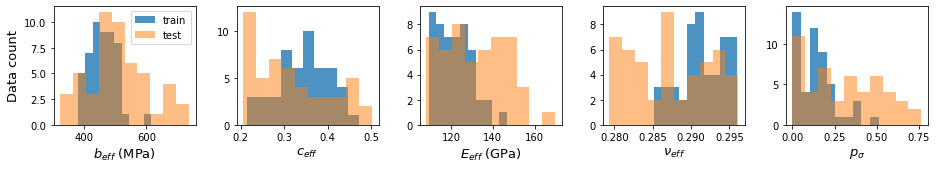}
\caption{}
\end{subfigure}
\caption{Materials surrogate modeling: (a) input data and (b) corresponding output data. The test set is generated by sampling input $\x$ from a uniform distribution, while the (second) training set is generated by sampling the input from a beta distribution, resulting in datasets that have different supports, useful to assess performance of algorithms for out-of-distribution prediction.\label{fig:materials_datasets}}
\end{figure}

Three datasets, two training and one test dataset, are generated by sampling the input $\x$ from different distributions: a uniform distribution within predefined bounds for the test and and first training set (used to study performance of the probabilistic data-driven model for in-distribution prediction), and a beta distribution for the second training set (to assess out-of-distribution performance). Fig. \ref{fig:materials_datasets} shows histograms of the test and second training datasets, both of size 50, illustrating the fact that outputs indeed have different supports. These datasets will allow a robust assessment of the epistemic uncertainty predictions of our novel BNN training scheme.

\begin{figure}[tb!]
\centering
\includegraphics[width=\textwidth]{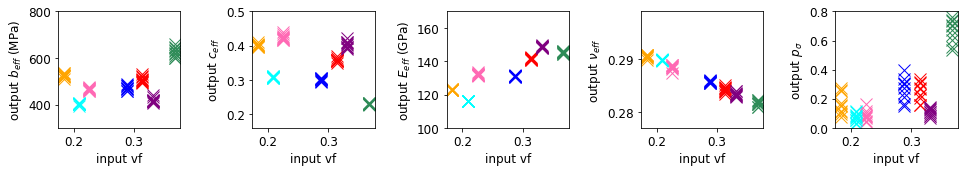}
\caption{Materials surrogate modeling: effect of random placement of fibers (aleatory uncertainty) on outputs $\y$.\label{fig:materials_aleatory}}
\end{figure}

By construction, the generated datasets are also embedded with aleatory uncertainties (noise), due to the random placement of fibers within the microstructure which yield different output values for a given input $\x$. Fig. \ref{fig:materials_aleatory} illustrates how this noise affects different outputs differently. For a given input $\x$, 10 microstructures were generated and the FE simulations were run to collect the corresponding outputs $\y_j$, $j=1:10$. This was performed for 7 different values of $\x$, represented by different colors in the scatter plots on \ref{fig:materials_aleatory}. For a given color, the scatter in the y-direction represents the uncertainty in the output solely due to noise. This aleatory uncertainty is very small for the effective Young's modulus $E_{eff}$, but quite large for the localized stress property $p_{\sigma}$, indicating that this output quantity is strongly affected by the random placement of fibers. These insights help design the aleatory noise model Eq. \eqref{eq:like}, in particular the covariance matrix $\Sigma_{\text{noise}}$, assumed known \text{a priori} in this work. In particular we assume a diagonal noise covariance matrix with following coefficients of variations for different outputs: $b_{eff}:3\%,c_{eff}:2\%,E_{eff}:0.5\%,\nu_{eff}:0.2\%,p_{\sigma}:25\%$.

\subsubsection{Functional prior construction}

In engineering applications, \textit{a priori} knowledge is often available in the form of low-fidelity models or empirical laws, and concepts of pre-training can help improve the generalization capabilities of deterministic neural networks, see e.g. \cite{Chen2023,Chen2024} for recent applications in mechanics or \cite{DE2022} which devises l1--regularization strategies to embed knowledge from low-fidelity data. Via ensembling with anchoring we frame concepts of pre-training within a Bayesian framework to guide both the mean but also the uncertainty predictions away from data. 

To design the functional prior density $p(g)$ in this non-trivial multi-input multi-output example, we first ran a simple sensitivity analysis on the training data to determine if some inputs had negligible influence on some of the outputs. Table \ref{tab:materials_mi} shows the mutual information between each input $\x[i]$ -- output $\y[j]$ pair, defined as:
$$I(\x[i],\y[j])=\int \int p(\x[i],\y[j])\log{\frac{p(\x[i],\y[j])}{p(\x[i])p(\y[j])}}, \quad i=1:4, j=1:5$$
The mutual information equates 0 only if input $\x[i]$ and output $\y[j]$ are independent, which is the case for some pairs of input-output in this example.

\begin{figure}[tb!]
\begin{subfigure}[b]{\textwidth}
\centering
\includegraphics[width=\textwidth]{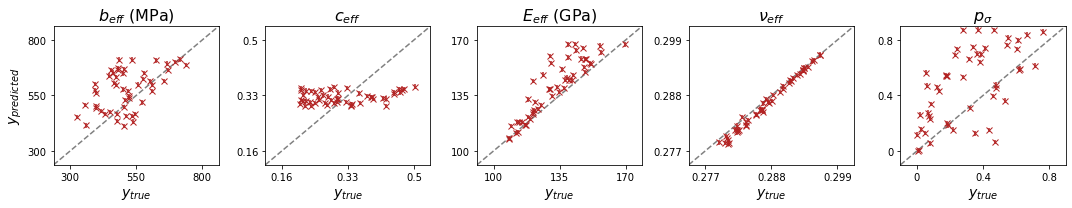}
\caption{}
\end{subfigure}
\begin{subfigure}[b]{\textwidth}
\centering
\includegraphics[width=0.9\textwidth]{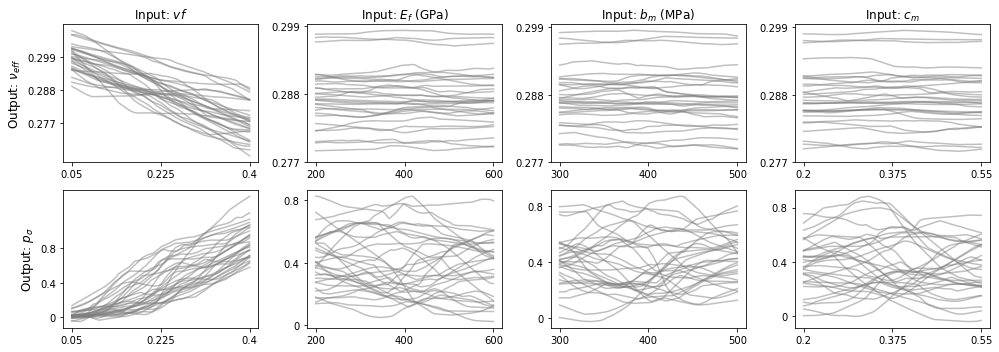}
\caption{}
\end{subfigure}
\caption{Materials surrogate modeling: prior design. (a) Performance of the prior mean function $m(\x)$ on the test set, x-axis represents the true output value from the data and the y-axis the prediction from the prior mean. (b) Sample functions $\kth{g}$ from the functional prior $p(g)$ for two specific outputs $\nu_{eff}, p_{\sigma}$, highlighting the non-zero mean trend as a function of $\text{vf}$, negligible effect of all other inputs on $\nu_{eff}$, and $p_{\sigma}>0$ constraint which affects the prior uncertainty for small values of the volume fraction.\label{fig:materials_prior}}
\end{figure}

We then we choose independent GP-based priors for each output $\y[j]$, which use as features only those inputs that exhibit some influence on the output (`influential subset' in Table \ref{tab:materials_mi}). The mean is chosen as a linear function of the form $m(\x) = \theta \text{vf} = \theta \x \left[ 0 \right]$, where $\theta$ is learnt via least-squares fit to the training data. This function is interesting as it is highly informative for some outputs (e.g. $\nu_{eff}$) and less so for others (e.g. $c_{eff}$), as illustrated in Fig. \ref{fig:materials_prior}~(a), so it allows a thorough study of the benefits of using an informative prior. With respect to the prior uncertainty, each GP prior uses a multivariate RBF kernel $0.2 \exp{\left(-\frac{\dist{\tilde{\x}-\tilde{\x}'}}{2 (0.8)^2} \right)}$ where $\tilde{\x}$ is the influential subset previously mentioned. Improved priors could leverage the sensitivity analysis to assign different lenghtscales for different inputs, but this analysis is left for future work. 

\begin{table}[tb!]
\centering
\begin{tabular}{ c || c | c | c | c | c }
Output $\y[j])$ & $b_{eff}$ & $c_{eff}$ & $E_{eff}$ & $\nu_{eff}$ & $p_{\sigma}$ \\
Input $\x[i]$ & & & & & \\ \hline \hline
$\text{vf}$ & 0.10 & 0.03 & 1.44 & 2.07 & 0.15 \\ \hline
$E_f$ & 0.06 & 0. & 0.07 & 0. & 0.05 \\ \hline
$b_m$ & 0.41 & 0.15 & 0. & 0. & 0.16 \\ \hline
$c_m$ & 0. & 1.47 & 0. & 0. & 0.12 \\ \hline \hline
Influential input subset & $\lbrace \text{vf},E_f,b_m \rbrace$ & $\lbrace \text{vf},b_m,c_m \rbrace$ & $\lbrace \text{vf},E_f \rbrace$ & $\lbrace \text{vf} \rbrace$ & $\lbrace \text{vf},E_f,b_m,c_m \rbrace$
\end{tabular}
\caption{Materials surrogate modeling: sensitivity analysis. Mutual information $I(\x[i],\y[j])$ (in nats) is computed using the scikit-learn package \cite{scikit-learn}.\label{tab:materials_mi}}
\end{table}

Finally, the prior density is also constrained within admissible bounds, namely $\y[j]>0$. This is performed by mapping sample functions $\kth{g}(\x)$ back to the admissible region ($>0$) whenever they violate the constraint. Fig. \ref{fig:materials_prior}~(b) displays some sample functions for two different outputs $\nu_{eff}$ and $p_{\sigma}$. With respect to the latter, the bottom left plot illustrates an instance where, for small values of the volume fraction input $\text{vf}\sim 0.05$, sample functions were often mapped back to the positive region, yielding a smaller posterior uncertainty in that region.

This analysis provided some useful guidelines on designing a functional prior for multi--input--output scenarios. Having defined this prior, we leverage our novel BNN training procedure to train a NN anchored ensemble to available training data, and assess the capabilities of this algorithm in both accuracy and quality of the uncertainty estimation.

\subsubsection{Performance metrics and algorithm validation for in-distribution data}

\begin{figure}[tb!]
\begin{subfigure}[b]{0.9\textwidth}
\centering
\includegraphics[width=\textwidth]{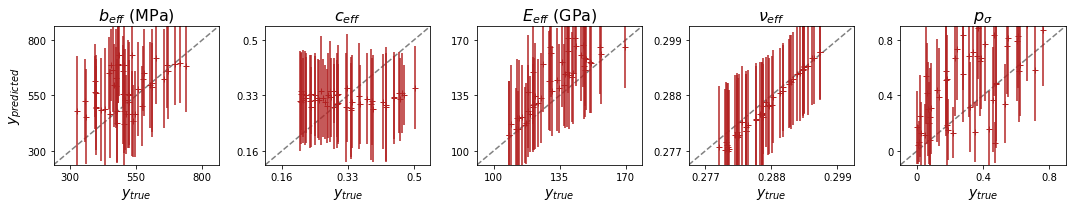}
\includegraphics[width=\textwidth]{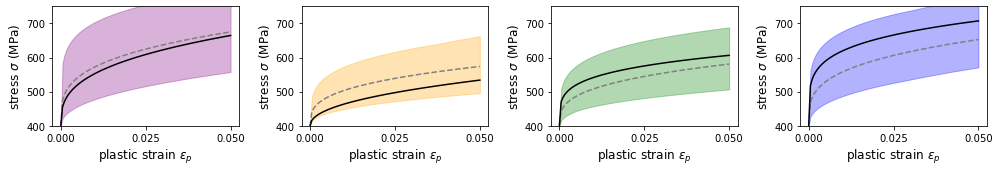}
\caption{}
\end{subfigure}
\begin{subfigure}[b]{0.9\textwidth}
\centering
\includegraphics[width=\textwidth]{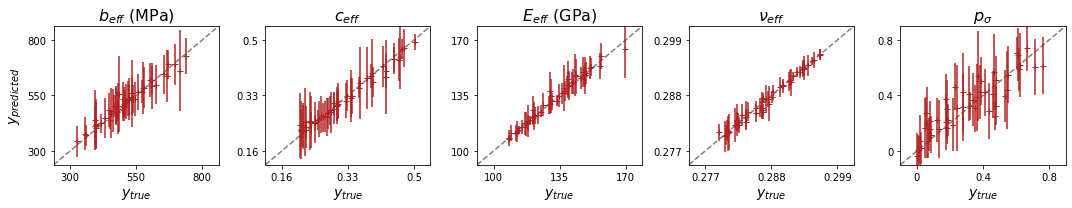}
\includegraphics[width=\textwidth]{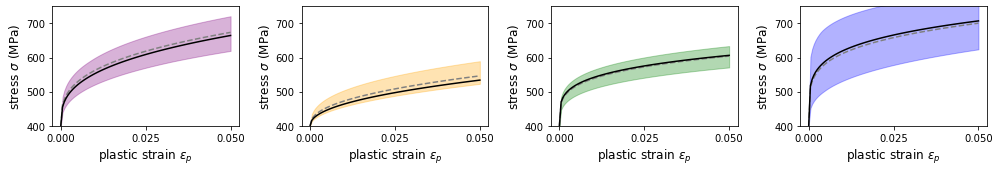}
\caption{}
\end{subfigure}
\begin{subfigure}[b]{0.9\textwidth}
\centering
\includegraphics[width=\textwidth]{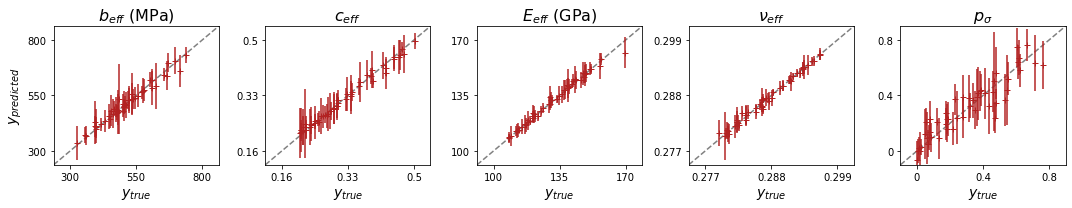}
\includegraphics[width=\textwidth]{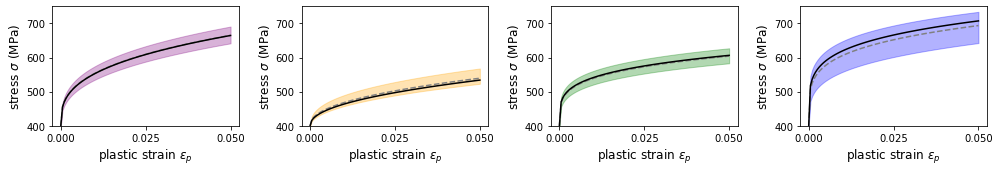}
\caption{}
\end{subfigure}
\caption{Materials surrogate modeling: qualitative assessment of our proposed BNN predictions for in-distribution test data when increasing the amount of training data; (a) prior, (b) 50 training data points and (c) 100 data points. In each subfigure, the top plot shows for each output the predicted mean $\pm$ 2 predicted standard deviations vs. the true value from data. The bottom plot shows the mean and uncertainty propagated to the effective behavior law $\sigma=a+b_{eff}(\varepsilon_{pl})^{c_{eff}}$ for four specific test points. \label{fig:materials_in_dist}}
\end{figure}

The NN architecture considered consists again in 4 hidden layers, 20 neurons in each layer with a leaky ReLU activation function. We first train our novel BNN (Alg. \ref{alg:ours}) on the first training dataset, sampled using the same distribution as the test set, using $N=25, 50, 100$ training data points. Qualitatively, Fig. \ref{fig:materials_in_dist} shows that predictions on the test set improve, and predicted epistemic uncertainties are reduced, as we increase the amount of training data. 

To perform a more robust quantitative assessment of the algorithm, we look at three metrics that aim to separately assess the accuracy of the mean prediction and quality of the uncertainty estimation. The root-mean squared error quantifies the accuracy of the mean prediction:
\begin{equation}
\text{RMSE} = \sqrt{\sum_{i=1}^{N_{test}} \left( \y_i - \expect{\y_i \vert \x_i, \mathcal{D}} \right)^2}
\end{equation}
where $\expect{\y_i \vert \x_i, \mathcal{D}}$ is computed as an average over the ensemble, Eq. \eqref{eq:predictive_moments_ens}. The calibration curve and miscalibration area \cite{Tran_2020} quantify the quality of the uncertainty estimation, by comparing the distribution of the predicted uncertainties with the distribution of the errors. A model is well calibrated if the calibration curve is close to the $x=y$ line, and miscalibration area is close to 0. Note that a model can be well calibrated even if it makes erroneous mean predictions, as long as the predicted uncertainties are also large. Finally, the expected log-pointwise predictive density \cite{Gelman2014} combines both mean and uncertainty assessment, and can be computed from the ensemble of NNs (with trained weights $\kth{\hat{\w}}$, $k=1:K$) as:
\begin{equation}
\text{ELPPD} = \sum_{i=1}^{N_{test}} \log{p(\y_i \vert \x_i, \mathcal{D})} = \sum_{i=1}^{N_{test}} \log{\left(\frac{1}{K} \sum_{k=1}^K p(\y_i \vert \x_i, \kth{\hat{\w}})\right)}
\end{equation}
Again, table \ref{tab:materials_results} shows that all metrics improve as the number of training data points increases.

\begin{table}[tb!]
\centering
\begin{tabular}{c || c | c | c | }
Metric & RMSE ($\downarrow$) & miscalibration ($\downarrow$) & ELPPD ($\uparrow$) \\ 
Training data / algorithm & & & \\ 
Prior ($N=0$) & 0.48 & 0.18 & - 4100 \\ \hline
InD ($N=25$) / anchored & 0.19 & 0.12 & - 196 \\ \hline
InD ($N=50$) / anchored & 0.13 & 0.11 & 33 \\ \hline
InD ($N=100$) / anchored & 0.10 & 0.09 & 155 \\ \hline \hline
OutD ($N=50$) / anchored & 0.16 & 0.09 & - 114 \\ \hline
OutD ($N=50$) / vanilla & 0.24 & 0.17 & - 459
\end{tabular}
\caption{Materials surrogate modeling: performance of BNNs. `InD' refers to in-distribution data, training and test sets come from the same distribution data, `OutD' refers to out-of-distribution data. `anchored' is the proposd algorithm. \label{tab:materials_results}}
\end{table}

\subsubsection{Performance for out-of-distribution predictions}

\begin{figure}[tb!]
\begin{subfigure}[b]{\textwidth}
\begin{center}
\includegraphics[width=\textwidth]{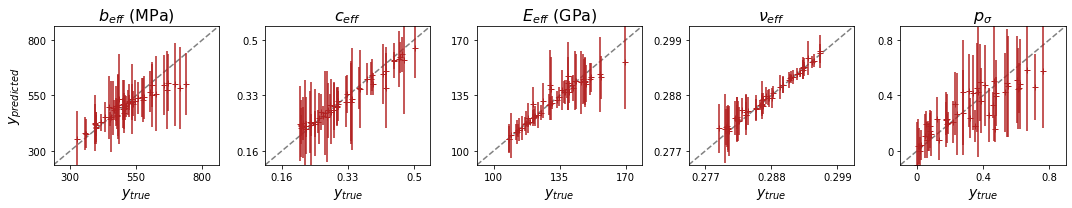}
\includegraphics[width=\textwidth]{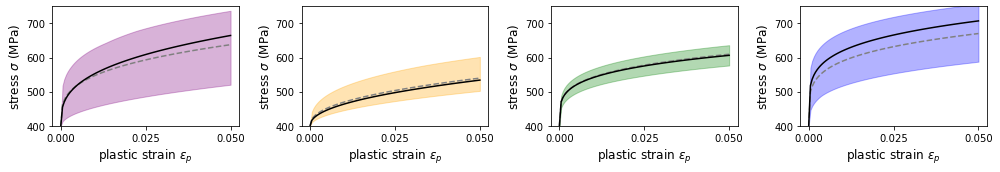}
\end{center}
\caption{}
\end{subfigure}
\begin{subfigure}[b]{\textwidth}
\centering
\includegraphics[width=\textwidth]{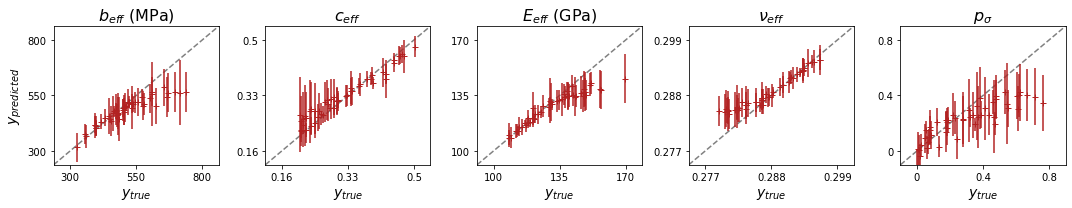}
\includegraphics[width=\textwidth]{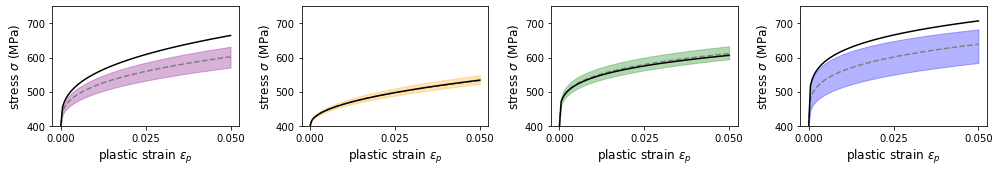}
\caption{}
\end{subfigure}
\caption{Materials surrogate modeling: qualitative assessment of a) our novel anchored BNN vs. b) vanilla ensembling for out-of-distribution predictions.\label{fig:materials_outD}}
\end{figure}

A more interesting scenario consists in training and testing on datasets that have different supports. Fig. \ref{fig:materials_outD}~(b) shows that in this case, vanilla ensembling tends to under-estimate uncertainties. This is very damaging since it provides the user with an incorrect sense of confidence in the predictions. BNN with anchoring, as proposed in Alg. \ref{alg:ours}, performs better with respect to both the mean, which is to be expected since it leverages some prior information, but also the uncertainty estimation, as illustrated qualitatively in Fig. \ref{fig:materials_outD} and quantitatively in Table \ref{tab:materials_results}. Also, in comparing performance for in-distribution and out-of-distribution predictions (for same number of training data points $N=50$) in Table \ref{tab:materials_results}, one can see that, although the mean prediction is less accurate for out-of-distribution data, which can be expected, the quality of the uncertainty estimation is comparable. This suggests that the algorithm is indeed capable of appropriately assessing confidence in its own predictions.

\begin{figure}[tb!]
\begin{center}
\includegraphics[width=.8\textwidth]{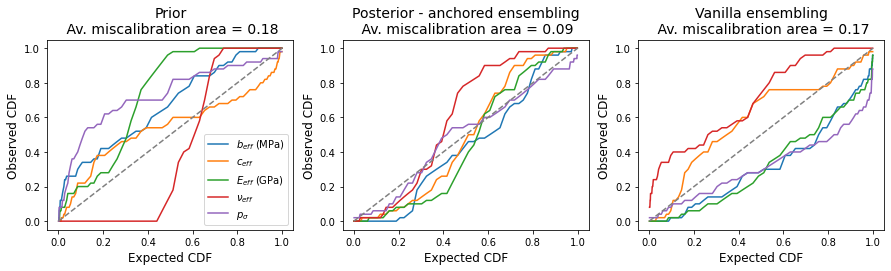}
\end{center}
\caption{Calibration curves per output for prior, anchored ensemble [proposed approach], and vanilla ensemble.\label{fig:materials_miscal}}
\end{figure}

Finally, while Table \ref{tab:materials_results} provides results averaged over all the outputs, Table \ref{tab:materials_results_2} shows the improvement (from vanilla to anchored ensembling) for the RMSE and miscalibration area for each output separately, evaluated as $\frac{\text{metric}_\text{vanilla} - \text{metric}_\text{anchored}}{\text{metric}_\text{vanilla}}$. Fig. \ref{fig:materials_miscal} compares the calibration curves. Our algorithm outperforms vanilla ensembling in terms of both mean accuracy and quality of the uncertainty prediction, except for a slight 2$\%$ increase in miscalibration area for $c_{eff}$. The outputs that benefit the most from prior knowledge in terms of mean accuracy are $\nu_{eff}$, $E_{eff}$, for which the prior mean was most informative. 

\begin{table}[tb!]
\centering
\begin{tabular}{c || c | c | c | c | c | }
Output $\y[j]$ & $b_{eff}$ & $c_{eff}$ & $E_{eff}$ & $\nu_{eff}$ & $p_{\sigma}$ \\ \hline
RMSE improvement ($\%$) & 0.22 & 0.23 & 0.39 & 0.66 & 0.27 \\ \hline
miscalibration improvement ($\%$) & 0.53 & -0.02 & 0.38 & 0.31 & 0.81 \\ \hline
\end{tabular}
\caption{Materials surrogate modeling: per--output accuracy improvement from vanilla to anchored ensembling (a positive number means an improvement).\label{tab:materials_results_2}}
\end{table}

\section{Conclusion}

The main contributions of this work are threefold. First, we performed a study on properties of NN parameter-space densities that are necessary to appropriately convey information in the function-space, highlighting in particular the importance of considering correlation between NN weights. Then, we derived an enhanced anchored ensembling procedure to perform approximate Bayesian inference in NNs, allowing the user to integrate knowledge from both the training data and a functional prior. This algorithm is scalable as it is easily parallelizable and relies on a low-rank approximation of the covariance between NN weights. Finally, we performed a robust assessment of this algorithm on a multi--input--output surrogate materials modeling task, evaluating performance with respect to both accuracy and quality of the uncertainty estimation, both of whom are important when leveraging data-driven models in mechanics and materials applications.

In this work we also provided some insights on how to design a functional prior. This task should be further elaborated, for instance by adapting recent strategies for NN architecture search to prior selection, such as \cite{SINGH2024} which leverages Bayesian model selection concepts, or \cite{MAULIK2023} which performs search within the ensembling framework. The algorithm also assumes knowledge of the aleatory noise model. As part of our ongoing work we are extending this algorithm to integrate noise covariance adaptation, using e.g. polynomial models as presented in \cite{Ozge2024} for heteroscedastic GPs. Finally, extensions of this algorithm to convolutional, recurrent or graph architectures would also prove interesting, building on such works as \cite{fortuin2022} which found that convolutional NNs in classification tasks display spatial correlations, again hinting at the importance of considering correlation between NN weights in ensembling methods for advanced architectures.

\bibliography{references} 

\begin{appendices}
\section{Details about pre-training to functional prior}\label{app:pretraining}

The proposed approach requires pre-training an ensemble of $K$ NNs to $K$ realizations of a functional prior $p(g)$. The pre-training of the ensemble is performed as detailed in Alg. \ref{alg:pre-training}. Note that throughout the manuscript we assume that the input has been scaled so that the region of interest for prediction is $\x \in \left[ -1, 1\right]$. Also, it is important to notice that the NNs in the ensemble are not randomly initialized -- instead they are initialized to a slightly perturbed version of a given set of weights $\w_{\text{init}}$. This is critical to appropriately capture the correlation between NN weights that is solely due to pre-training rather than random initialization. Finally, in this pre-training stage the goal is to generate an ensemble of NNs that perfectly represents the functional prior, thus there is no regularization at this stage and no issue with over-fitting. The optimization is run for as many epochs as it takes to accurately match the prior dataset, which depends on the complexity of the prior functions. For instance, in the prior study, fitting to prior A, B and C took 100, 200 and 2,000 epochs respectively.

\begin{algorithm*}[ht!]
\begin{algorithmic}
\State{\textbf{Input}: Functional prior $p(g)$, $g: \x \rightarrow \y$}
\State{\textbf{Output}: Ensemble of NNs with pre-trained weights $\kth{\w}$, $k=1:K$ that match the prior}
\State
\State Sample prior input set $\x_{j}$, $j=1:M$ from a standard Gaussian using large $M$ (e.g. 500) and Latin Hypercube Sampling to ensure appropriate coverage of the input space.
\State Generate one set of weights $\w_{\text{init}}$ via e.g. He initialization.
\For {$k=1:K$}
\State Sample realization from functional prior $\kth{g} \sim p(g)$
\State Generate prior dataset: $\lbrace \x_j, \y_j=\kth{g}(\x_j)\rbrace_{j=1:M}$
\State Initialize NN weights to a slightly perturbed version of $\w_{\text{init}}$, e.g., $\w_{\text{init}} + 0.01 \varepsilon, \quad \varepsilon \sim \gpdf{0}{\mathbb{I}_d}$
\State Fit NN weights to prior dataset: 
$$\kth{\w} = \text{argmin} \quad \sum_{j=1}^{M} \dist{\y_j - \nn{\w}{\x_j}} $$
\EndFor
\end{algorithmic}
\caption{Ensemble pre-training}\label{alg:pre-training}
\end{algorithm*}

\section{SVD and correlation matrix}\label{app:svd}

Consider a centered data matrix $\mathbb{W}$ of size $K \times d$ so that $K << d$. Its SVD is given as $\mathbb{W}=U S V^T$ where $U$ is $K \times K$, $S$ has $K$ non-zero singular values and $V$ is $d \times K$ so that $V^T V = \mathbb{I}_{K}$. Then the covariance matrix of the data is given by:
$$ \Sigma = \frac{1}{K-1} \mathbb{W}^T \mathbb{W} = \frac{1}{K-1} VSU^T USV^T = \frac{1}{K-1} VS^2V^T $$
The generalized inverse of $\Sigma$, defined so that $\Sigma \Sigma^{+} \Sigma=\Sigma$ is given as $\Sigma^{+}=(K-1)VS^{-2}V^T$ since:
$$
\Sigma \Sigma^{+} \Sigma = \frac{1}{K-1} VS^2V^T (K-1)VS^{-2}V^T \frac{1}{K-1} VS^2V^T = \frac{1}{K-1} V S^2V^T = \Sigma
$$

Also, sampling from the corresponding degenerate multivariate Gaussian density with mean $\mu$ and low-rank covariance $\Sigma$, $p(\w)=\gpdfX{\w}{\mu}{\Sigma}$ can be performed efficiently by sampling standard Gaussian random variates $z$ in dimension $K << d$, and transforming them as $\w=\frac{1}{\sqrt{K-1}} V S z + \mu$.

\end{appendices}

\end{document}